\documentclass[10pt]{article}
\usepackage[T1]{fontenc}
\usepackage[utf8]{inputenc}
\usepackage[sc]{mathpazo}

\usepackage[letterpaper,left=.75in,right=.75in,top=.7in,bottom=.65in,headheight=12pt,headsep=10pt]{geometry}
\usepackage{amsmath,amssymb,mathtools}
\usepackage{booktabs,array,tabularx,longtable}
\usepackage{graphicx}
\usepackage{xcolor}
\definecolor{burgundy}{HTML}{762D45}
\usepackage[font=small,labelfont=bf,labelsep=period,justification=justified,singlelinecheck=false]{caption}
\usepackage{subcaption}
\usepackage{placeins,dblfloatfix,balance,needspace}
\usepackage{enumitem,microtype,xcolor}
\usepackage[numbers,sort&compress]{natbib}
\usepackage{xurl,hyperref,bookmark}
\definecolor{linkblue}{HTML}{762D45}
\definecolor{abstractbg}{HTML}{F6EFF0}
\hypersetup{colorlinks=true,linkcolor=linkblue,citecolor=linkblue,urlcolor=linkblue,pdftitle={PBit1: One Programmable P-Bit per Parameter — A Finite-Control Representation of Pretrained Networks},pdfauthor={Omar Ramadan}}
\usepackage{fancyhdr,titlesec}
\setlist{nosep,leftmargin=*}
\titleformat{\section}{\large\bfseries\color{burgundy}}{\thesection}{.65em}{}
\titlespacing*{\section}{0pt}{10pt plus 2pt minus 1pt}{4pt}
\titleformat{\subsection}{\normalsize\bfseries\color{burgundy}}{\thesubsection}{.6em}{}
\titlespacing*{\subsection}{0pt}{7pt plus 1pt}{3pt}
\titleformat{\subsubsection}{\normalsize\bfseries\color{burgundy}}{\thesubsubsection}{.6em}{}
\titlespacing*{\subsubsection}{0pt}{6pt plus 1pt}{3pt}
\titleformat{\paragraph}[runin]{\normalsize\bfseries}{}{0pt}{}[.]
\titlespacing*{\paragraph}{0pt}{5pt}{.5em}

\newcommand{\Var}{\operatorname{Var}}

\newcommand{\Bern}{\operatorname{Bernoulli}}
\newcommand{\Bin}{\operatorname{Binomial}}
\newcommand{\MSE}{\operatorname{MSE}}

\newcommand{\sigmoid}{\operatorname{sigmoid}}
\newcommand{\clip}{\operatorname{clip}}

\newcommand{\code}[1]{{\small\nolinkurl{#1}}}
\newcommand{\artifact}[1]{\href{https://github.com/oramada/pbit1-research/tree/main/results/thrml_spw/spark_depth_20261106/#1}{\nolinkurl{#1}}}
\newcommand{\figfull}[4]{\begin{figure}[t]\centering\includegraphics[width=#2\linewidth]{figures/#1}\caption{#3}\label{#4}\end{figure}}

\begin{document}
\thispagestyle{plain}
\begin{center}
{\fontsize{18}{21}\selectfont\bfseries\color{burgundy} PBit1: One Programmable P-Bit per Parameter\\[2pt]
A Finite-Control Representation of Pretrained Networks\par}
\vspace{12pt}
{\normalsize Omar Ramadan\par}
\vspace{3pt}
{\small Johns Hopkins Whiting School of Engineering\par}\vspace{2pt}
{\small \href{mailto:oramada1@jh.edu}{oramada1@jh.edu}\par}
\vspace{6pt}
{\small\href{https://github.com/oramada/pbit1}{Code}\quad
\href{https://github.com/oramada/pbit1-research}{Research archive}}
\end{center}
\vspace{5pt}
\noindent\colorbox{abstractbg}{\begin{minipage}{\dimexpr\textwidth-2\fboxsep\relax}
\vspace{5pt}\normalsize
We introduce PBit1, a finite-control representation of pretrained neural networks that assigns one programmable p-bit to each learned parameter. A 12-bit command sets its probability, a shared affine range maps that probability to the weight domain, and temporal observations reconstruct the weight for inference. Operator-aware range grouping and sensitivity-based sampling allocation adapt the representation to the network. Across five independent weight draws at 32,768 observations per scalar, selected configurations preserve task quality in GPT-2, DistilBERT, and LFM3-1.7B with every learned scalar converted and no fine-tuning. GPT-2 and LFM3 perplexity increase by approximately 0.66\% and 0.26\%, respectively, while DistilBERT accuracy changes from 87.51\% to 87.49\%. A sensitivity-based policy reduces GPT-2 reconstruction observations by 41.7\% while retaining the quality criterion. Complete-model evaluations reuse reconstructed weights in BF16 inference, and THRML operator experiments examine native sampling and spatial--temporal tradeoffs. Together, these results establish PBit1 as a concrete construction for studying finite-control weight representation and temporal reconstruction in pretrained networks.

\vspace{7pt}
\begin{center}\includegraphics[width=\linewidth]{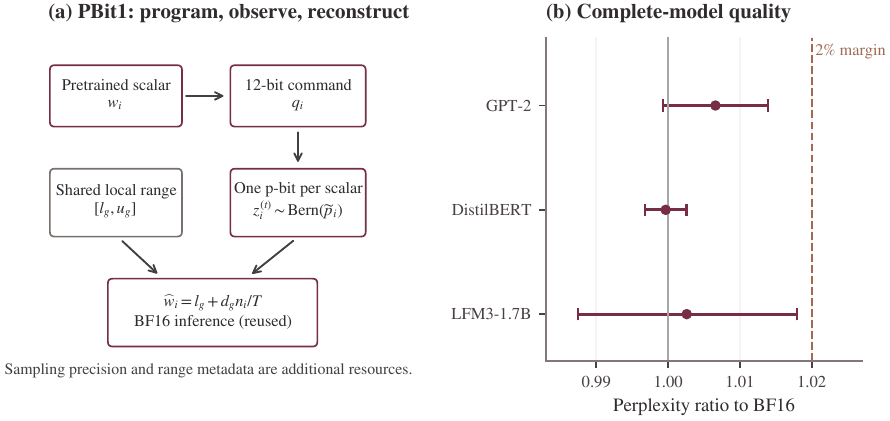}\end{center}
\vspace{-8pt}
\captionof{figure}{\textbf{PBit1 separates stochastic-element count from reconstruction precision.} The left panel shows how finite probability commands program one logical p-bit per learned scalar. Shared ranges and temporal counts reconstruct weights for reuse in BF16 inference. The right panel shows that selected configurations meet the predefined 2\% perplexity-ratio criterion on three complete models at $T=32{,}768$. Points and intervals show five-draw means in log-loss space and conditional 95\% $t$ intervals. DistilBERT uses categorical perplexity and also meets its accuracy criterion.}\label{fig:quality}
\vspace{5pt}
\end{minipage}}
\clearpage
\twocolumn

\section{Introduction}
A programmable probabilistic bit has a binary instantaneous state and a tunable stationary probability. Repeated observations estimate that probability, allowing additional time to improve precision with the same stochastic-element count. This suggests a representation for pretrained networks in which each learned parameter is assigned one logical programmable p-bit. The central question is how faithfully a network can be reconstructed from these elements under finite probability controls and sampling budgets.

We introduce \textbf{PBit1}, which combines one programmable p-bit per learned parameter with shared affine ranges and temporal reconstruction. Encoding sets the probability command, and decoding maps an observed count back to a weight. The shared range determines the estimation scale, while the sampling policy determines the temporal budget. Together, these components convert an existing checkpoint into reconstructed weights without fine-tuning. We denote the one-element-per-scalar assignment by K1.

PBit1 builds on probability-based representations and progressive stochastic computation \citep{shawe1995,hartmann2019}. Its finite-control construction covers every learned scalar and combines operator-aware shared ranges with sensitivity-based sampling. We evaluate how these choices affect task fidelity, storage, and the number of observations required for reconstruction.

We examine three research questions.
\begin{enumerate}[leftmargin=*,itemsep=2pt]
\item Does the representation preserve the pretrained function? Complete-model evaluations test task quality on GPT-2, DistilBERT, and LFM3-1.7B with every learned scalar converted.
\item What affects the required sampling budget? Range grouping and sensitivity-based allocation determine how reconstruction work is distributed across the network.
\item What resources and tradeoffs does the representation introduce? Storage and sampling costs are evaluated alongside native THRML comparisons of one-element and multi-element representations.
\end{enumerate}
The GPT-2 layout study connects these questions by showing how range grouping changes reconstruction quality at a fixed sampling budget.

Complete-model evaluations reconstruct weights once and reuse them in BF16 inference. This makes it possible to study the representation at network scale while controlling the reconstruction process. Direct decoding of the programmed mean provides a reference for the error introduced by temporal sampling, and native THRML experiments examine the corresponding behavior at operator scale.

\section{Background and Related Work}
\paragraph{Stochastic representations}
Stochastic computing represents numerical values through the statistics of bit streams, with arithmetic and accuracy shaped by stream length and correlation \citep{alaghi2013}. Bernoulli bit-stream representations also have a long history in neural computation \citep{shawe1995}. Probabilistic binary networks use stochastic weight distributions and ensembles \citep{peters2018}, while progressive stochastic binarization supports pretrained-network conversion, precision through accumulation, and adaptive computational allocation \citep{hartmann2019}. PBit1 brings these ideas into a specified finite-control construction with one programmable stochastic element per pretrained scalar, shared affine ranges, and temporal weight reconstruction. Its evaluation follows the complete conversion of three pretrained checkpoints and the effect of range and sampling policies on their task quality.

\paragraph{Post-training quantization}
GPTQ uses post-training reconstruction for transformer quantization \citep{frantar2023}, and AWQ uses activations to protect important weight directions \citep{lin2024}. AdaRound adapts rounding decisions to a data-dependent reconstruction objective \citep{nagel2020}, while SmoothQuant redistributes quantization difficulty between activations and weights through an equivalent scaling transformation \citep{xiao2023}. PBit1's grouping policy also uses activation-dependent sensitivity. Temporal sampling introduces a variance term and an observation budget in addition to the encoding approximation, linking the choice of local range to the time required for reconstruction.

\paragraph{Probabilistic hardware}
Programmable p-bits provide tunable stochastic binary units for probabilistic circuits \citep{camsari2017}. Magnetic-tunnel-junction implementations have demonstrated connected p-bit networks for integer factorization \citep{borders2019}. Extropic's Z1T work studies sparse transformer-like computation with multi-p-bit representations and temporally accumulated statistics \citep{extropic2026z1t}. We use the official THRML sampler \citep{thrml} to compare one-element and four-element constructions on a pretrained operator bank. These experiments complement the full-model reconstruction study with direct measurements of native sampling behavior.

\section{Method}
\subsection{Affine stationary-probability encoding}
For parameter $i$ in group $g(i)$, let $c_g$ be the center and $r_g\geq0$ the radius. The ideal representation is
\begin{equation}
 w_i=c_g+r_g(2p_i-1),\qquad z_i^{(t)}\sim\Bern(p_i).
 \label{eq:encoding}
\end{equation}
With independent samples, the temporal estimate is
\begin{equation}
 \begin{aligned}
 \hat w_{i,T}&=c_g+r_g\left(\frac{2}{T}\sum_{t=1}^T z_i^{(t)}-1\right),\\
 \Var(\hat w_{i,T})&=\frac{4r_g^2p_i(1-p_i)}{T}.
 \end{aligned}
 \label{eq:variance}
\end{equation}
For $r_g=1$, the worst-case standard deviation is $T^{-1/2}$. The corresponding error-based precision measure, $b_{\rm eff}=-\log_2\sigma$, grows as $\tfrac12\log_2T$. Here, effective precision describes the sampling error of the scalar estimate.

\subsection{Finite probability programming}
The implementation stores the group minimum $l_g$ and span $d_g=u_g-l_g$ as FP32 values. This is algebraically equivalent to $c_g=l_g+d_g/2$ and $r_g=d_g/2$, up to floating-point evaluation. For a nonconstant group,
\begin{align}
 p_i^*&=(w_i-l_g)/d_g,\quad \epsilon=\sigmoid(-8),\\
 h_i^*&=\tfrac12\operatorname{logit}(\clip(p_i^*,\epsilon,1-\epsilon)),\\
 q_i&=\operatorname{round}\left[\frac{4095}{8}(h_i^*+4)\right],\\
 q_i&\in\{0,\ldots,4095\},\\
 \tilde p_i&=\sigmoid(2\hat h_i),\quad\hat h_i\simeq8q_i/4095-4.
 \label{eq:controller}
\end{align}
Decoding evaluates the field in FP32 and the sigmoid in FP64, with operation order fixed in the source. Constant groups use $p_i^*=1/2$ and $d_g=0$, so sampling does not change their value. For nonconstant groups, finite fields and quantized controls introduce bias, including at the endpoints. Increasing $T$ reduces variance around the programmed mean $l_g+d_g\tilde p_i$, with the encoding bias set by the finite command.

For a deterministic target and independent samples,
\begin{equation}
 \begin{aligned}
 \MSE(\hat w_{i,T},w_i)&=\left(l_g+d_g\tilde p_i-w_i\right)^2\\
 &\quad+\frac{d_g^2\tilde p_i(1-\tilde p_i)}{T}.
 \end{aligned}
 \label{eq:biasvariance}
\end{equation}
The finite commands and shared ranges are stored alongside the stochastic states. Section~\ref{sec:resources} accounts for these components.

\subsection{Temporal readout and execution}
We compare two execution regimes. \textbf{R1} reconstructs sampled weights before deterministic inference, and \textbf{R2} accumulates linear-operator outputs with sampled weights. All complete-model results use R1. We draw $n_i\sim\Bin(T,\tilde p_i)$ and reconstruct $l_g+d_gn_i/T$, which has the same marginal distribution as averaging $T$ independent Bernoulli observations. This avoids generating each observation explicitly.

For fixed inputs and identical weight samples, linearity gives
\begin{equation}
 X\left(\frac1T\sum_t W^{(t)}\right)=\frac1T\sum_t XW^{(t)}.
 \label{eq:linearity}
\end{equation}
Native operator tests verify this identity numerically. For fixed inputs and identical samples, R1 and R2 therefore have the same linear output and sampling variance. In R1, the reconstructed matrix is reused across the dataset, so examples share the same weight draw. Nonlinear operations remain outside this averaging identity.

\subsection{Range and sampling policies}
A PBit1 configuration consists of a fixed grouping of the pretrained scalars, their finite probability commands, shared minima and spans, and a temporal budget for each tensor. Changing a group's membership changes its span and the resulting reconstruction error. Changing its sample budget changes the variance without changing its programmed mean. Both choices configure the representation around the pretrained weights. The experiments below evaluate operator-aware grouping and a gradient-based temporal allocation as configuration policies for this mechanism.

\section{Experimental Setup}
\subsection{Models, data, and numerical execution}
Table~\ref{tab:models} lists the evaluated checkpoints and tasks. GPT-2 \citep{radford2019} and LFM3 \citep{liquid2026} are evaluated on WikiText-2 \citep{merity2016}, using their respective tokenizers and fixed article/window layouts. The SST-2-finetuned DistilBERT checkpoint \citep{sanh2019} is evaluated on IMDb \citep{maas2011}; its BF16 reference uses the same transfer between tasks. We convert every learned scalar, including biases and normalization parameters. Tied input embeddings and output heads use one shared representation and are counted once.

\begin{table*}[t]\centering\small
\caption{Complete evaluated checkpoints and task coverage. Token counts denote scored prediction targets.}\label{tab:models}
\begin{tabular}{lrrl}\toprule
Model & Unique parameters & Scored cases & Task\\\midrule
GPT-2 & 124,439,808 & 283,225 tokens & WikiText-2\\
DistilBERT & 66,955,010 & 25,000 reviews & IMDb classification\\
LFM3-1.7B & 1,728,801,536 & 294,442 tokens & WikiText-2\\\bottomrule
\end{tabular}
\end{table*}

Full-model inference uses BF16 weights and computation with deterministic settings. GPT-2 scores 586 windows across 62 articles; LFM3 scores 607 windows across the same articles. DistilBERT truncates inputs at 256 tokens and uses batches of 32. Runs execute locally on an NVIDIA DGX Spark/GB10; LFM3 uses CPU master weights and GPU offloading. The supplement specifies the runtime, loss accumulation, checkpoint revisions, and execution details.

\subsection{Selection and statistical protocol}
We use training activations to choose which matrix direction is grouped for each operator. Groups contain 128 consecutive entries after the matrix is flattened in the chosen direction; LFM3 uses groups of seven for depthwise convolution filters and sixteen for normalization tensors. The uniform policy uses 12-bit controls and $T=32{,}768$ samples per scalar. Each policy and seed list is fixed before its complete evaluation. The benchmarks had been inspected during earlier experiments, and the reported evaluations use fixed policies and seed lists on these datasets.

Let $\ell_0$ be the mean BF16 negative log likelihood and $\ell_s$ the value for independent weight draw $s$. For five seeds, $d_s=\ell_s-\ell_0$ and
\begin{equation}
 \begin{aligned}
 R&=\exp(\bar d),\\
 I_{95}&=\exp\left[\bar d\ \pm\ 2.776445\frac{s_d}{\sqrt5}\right].
 \end{aligned}
 \label{eq:ci}
\end{equation}
The predefined perplexity criterion requires the upper endpoint of $I_{95}$ to be at most $1.02$. This interval measures Monte Carlo variability conditional on the fixed evaluation data. DistilBERT must also have an upper 95\% bound on absolute accuracy decline of at most two percentage points. For DistilBERT, categorical perplexity is the exponential of classification cross-entropy. All five prespecified seeds enter the reported statistics.

We independently rescore saved per-example or per-token outputs and use a separate SDK to reconstruct weights and replay prespecified neural windows. These checks test implementation reproducibility. The supplement also reports article/seed bootstrap intervals to characterize variability across the evaluated articles.

\section{Results}
\subsection{Complete-model reconstruction feasibility}
\begin{table*}[t]\centering\small
\caption{Five-seed complete-model evaluations at $T=32{,}768$. Perplexities for stochastic language models are exponentials of mean seed NLL. Brackets give conditional Monte Carlo 95\% $t_4$ intervals for the ratio.}\label{tab:results}
\begin{tabular}{lrrl}\toprule
Condition & BF16 PPL & K1 PPL & PPL ratio [95\% interval]\\\midrule
GPT-2, original layout & 25.50149 & 27.02231 & 1.05964 [1.04254, 1.07701]\\
GPT-2, selected layout & 25.50149 & 25.66953 & 1.00659 [0.99927, 1.01397]\\
LFM3, selected layout & 18.93015 & 18.97918 & 1.00259 [0.98751, 1.01790]\\\bottomrule
\end{tabular}
\end{table*}

The selected GPT-2 and LFM3 layouts meet the predefined 2\% criterion (Table~\ref{tab:results}). DistilBERT also meets both criteria, with mean accuracy of 87.49\% against 87.51\% for BF16 and an upper interval endpoint of 0.084 percentage points for the decline. Its categorical-perplexity ratio is 0.99966 [0.99676, 1.00256]. GPT-2's original layout exceeds the quality margin at the same sample count, motivating the range-grouping analysis below.

\paragraph{Deterministic decoding reference}
Directly decoding the stored commands gives the finite-controller mean $\mu_i=l_g+d_g\tilde p_i$. Table~\ref{tab:mean} compares this model with BF16 and sampled reconstruction. The controller mean captures the encoding approximation, while the sampled result also includes temporal estimation error and BF16 rounding. For LFM3, the supplementary joint seed/article bootstrap gives a sampled ratio interval of [0.99188, 1.01065].

\begin{table}[t]\centering\small
\caption{Perplexity under deterministic decoding and sampled reconstruction using selected layouts. Sampled values exponentiate mean five-seed NLL; deterministic values are single decoded models.}\label{tab:mean}
\begin{tabular}{lrr}\toprule
Condition & GPT-2 & LFM3\\\midrule
BF16 reference & 25.50149 & 18.93015\\
Controller mean & 25.47279 & 18.95199\\
Sampled, $T=32{,}768$ & 25.66953 & 18.97918\\\bottomrule
\end{tabular}
\end{table}

\Needspace{7\baselineskip}
\subsection{Configuring reconstruction error}
\subsubsection{Operator-aware range grouping}
For a fixed input $x$ and independently sampled weights, the variance of output coordinate $j$ is
\begin{equation}
 \Var(\hat y_j\mid x)=\sum_i x_i^2\frac{d_{ji}^2\tilde p_{ji}(1-\tilde p_{ji})}{T}.
 \label{eq:outvar}
\end{equation}
Equation~\ref{eq:outvar} shows how range grouping affects the operator. Reconstruction variance is weighted by the input activations, so the same weight error can have different effects in different parts of the network. Row-wise and column-wise grouping assign different sets of weights to each range, changing both the programmed means and sampling variances. Orientation selection uses this structure to reduce error in sensitive directions at a fixed element count.

In GPT-2, changing the range orientation brings the reconstruction within the quality margin at $T=32{,}768$ (Table~\ref{tab:results}). The output use of the tied embedding matrix is especially sensitive. On a fixed training diagnostic, restoring that use to BF16 reduces PPL from 58.98 to 46.10, near the 45.80 reference. Restoring the input use alone gives 59.66. Reorienting the shared stochastic matrix gives 45.27 in the same draw, while retaining a single representation for both uses. With clean hidden activations held fixed, the predicted head-logit variance falls from 0.614 to 0.00270, a reduction of approximately 228-fold.

The preferred orientation varies across operators. The same reorientation increases predicted variance by approximately 3.2 times in one examined DistilBERT feed-forward matrix. LFM3 also shows an interaction between local and network-wide sensitivity. The block that is most harmful when substituted into the clean network differs from the block whose restoration helps the fully stochastic network most. The supplement develops these observations through interventions, signal-scale analysis, and conditional output-error covariance.

\subsubsection{Sensitivity-based temporal allocation}
Once a suitable range layout is chosen, sampling work can also be redistributed. Uniform sampling spends the same number of observations on parameters with different sensitivities. We instead use a first-order proxy that weights reconstruction variance by squared loss gradients. With variance coefficient $A_g$ and $N_g$ parameters in group $g$, the continuous allocation problem is
\begin{equation}
 \min_{T_g>0}\sum_g N_gT_g\quad\text{subject to}\quad
 \sum_g A_g/T_g\leq V
\end{equation}
whose solution has $T_g\propto\sqrt{A_g/N_g}$ before bounds and rounding. This relationship defines a first-order sampling heuristic. The implemented policy chooses orientations and sample counts from training/calibration data, with groups of sixteen for small normalization and bias vectors. The supplement gives the discrete allocation procedure used with the fixed pretrained weights.

A fresh five-seed GPT-2 evaluation scores all 283,225 targets. The selected policy reduces nominal observations per reconstruction from 4,077,643,628,544 to 2,378,245,865,472, a reduction of 41.7\%, while adding 53,088 bytes of range metadata. Every one of the 124,439,808 parameters retains a K1 assignment, and no sample count exceeds 32,768. The PPL ratio is 1.00780 [0.99907, 1.01660], meeting the same quality criterion (Figure~\ref{fig:allocation}).

\figfull{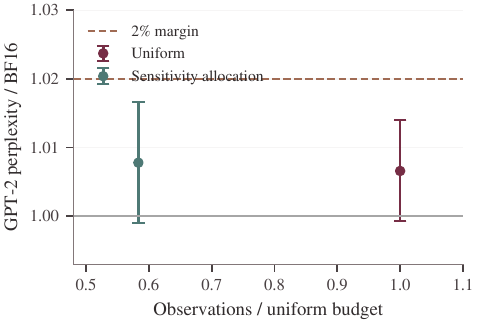}{1}{Validated sensitivity-based allocation reduces nominal reconstruction work to 58.324\% of the uniform budget. Vertical bars show independent five-seed Monte Carlo intervals. The sampled matrix is reused across inputs.}{fig:allocation}

\subsection{Spatial tradeoffs and nonideal controls}
\subsubsection{A stronger spatial baseline}
We examine the tradeoff between stochastic-element count and reconstruction error using THRML on an $8\times768$ GPT-2 output-head bank and eight fixed hidden vectors. The K4 control encodes normalized $u$ using three clamped coarse bits, $q=\min(\lfloor8u\rfloor,7)$, and one probabilistic fractional bit with target $f=8u-q$. Decoding gives $(q+z)/8$. The fractional command uses the same 12-bit field grid as K1.

\begin{table*}[t]\centering\small
\caption{Native THRML reoriented-head control, five draws. Scheduled observations count all nodes; free updates count only stochastic nodes. Output MSE is averaged over the fixed operator inputs.}\label{tab:native}
\begin{tabular}{lrrrr}\toprule
Condition & Elements & $T$ & All-node observations & Mean MSE\\\midrule
K1 & 6,144 & 32,768 & 201,326,592 & $7.0041\!\times\!10^{-5}$\\
K4 coarse/fractional & 24,576 & 8,192 & 201,326,592 & $3.3115\!\times\!10^{-6}$\\
K4 coarse/fractional & 24,576 & 32,768 & 805,306,368 & $5.6909\!\times\!10^{-7}$\\\bottomrule
\end{tabular}
\end{table*}

At matched scheduled observations, K4 has lower operator error than K1 (Table~\ref{tab:native}), at the cost of more elements and coarse information. Scheduled observations include clamped nodes; free stochastic updates differ. The three rows use 201,326,592, 50,331,648, and 201,326,592 free updates, respectively. Counting controls and instantaneous states gives 13 bits per K1 scalar and 19 per K4 scalar before shared ranges. Clamped-node reads, holding, and programming complete the corresponding hardware cost model.

\figfull{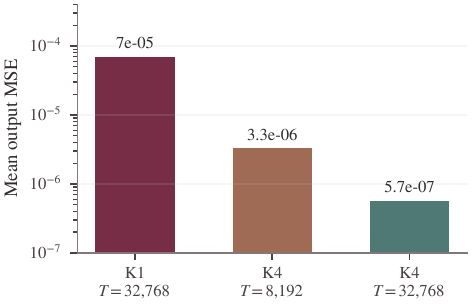}{1}{Native THRML comparison for the reoriented output-head bank. Bars show empirical five-draw mean output MSE, without confidence intervals. K4 at $T=8{,}192$ and K1 at $T=32{,}768$ have equal scheduled all-node observations; their free-node update counts differ (Table~\ref{tab:native}).}{fig:native}

\subsubsection{Noise distribution and device mismatch}
Gaussian perturbations matched to the finite-controller mean and estimator variance give similar GPT-2 training-diagnostic quality to Bernoulli reconstruction, with five-draw PPL ratios are 1.00406 [0.98855, 1.01981] and 0.99688 [0.98409, 1.00984], respectively. The matched-noise comparison relates the observed task quality to the scale of reconstruction error.

Second, simulated device gain and offset increase GPT-2 PPL relative to ideal reconstruction, with a paired ratio of 1.03229 [1.00038, 1.06522] on 1,024 training targets. An oracle correction using exact device parameters reduces mean squared bias by about 447-fold. Storing two FP32 calibration values per GPT-2 scalar would add approximately 996 MB. The supplement gives the mismatch model, calibration controls, and simulations of correlated samples and drift.

\section{Resource Accounting and Implementation}\label{sec:resources}
The K1 representation stores probability commands and shared ranges alongside one stochastic state per parameter. For $N$ parameters and $G$ groups, packed commands and FP32 ranges require $12N+64G$ bits. Including one instantaneous state per parameter gives $13N+64G$ bits. The current implementation stores commands in uint16 arrays, requiring $16N+64G$ bits before states. At group size 128, the totals including states are 13.5 bits per parameter with packed controls and 17.5 with uint16 controls. Reconstructed BF16 weights and framework memory are additional.

For LFM3, packed commands and ranges would occupy 2,701,541,888 bytes; the implemented uint16 representation occupies 3,565,942,656 bytes. R1 also holds 3,457,603,072 bytes of reconstructed BF16 weights, and retaining the reference model adds another copy. With weight offloading, the maximum recorded GPU allocation is 824,188,928 bytes, in addition to the host-resident data.

A hardware implementation would distribute energy across switching, bias programming, retention, readout, accumulation, and recalibration. Capacity and correlation-adjusted sampling rate would determine how the logical representation is scheduled. If programming is reused over $M$ inferences, the per-inference energy can be organized as
\begin{equation}
 \begin{aligned}
 E_{\rm infer}&=E_{\rm transitions}+E_{\rm readout}+E_{\rm accumulation}\\
 &\quad+\frac{E_{\rm program}}{M}+E_{\rm calibration}+E_{\rm retention}.
 \end{aligned}
\end{equation}
R1 amortizes reconstruction across inputs. A native R2 implementation would instead accumulate sampled operator outputs during execution, giving a different schedule for readout and accumulation. The terms above provide a cost model for evaluating those implementations.

\paragraph{Reproducibility}
Saved-output rescoring checks the evaluation results, and an independent SDK reproduces weight reconstruction and selected neural windows or batches. These prespecified replays match bitwise on the recorded runtime for GPT-2, DistilBERT, and LFM3. The supplement records their coverage, native checks, and evidence identifiers. The SDK provides finite-controller reconstruction with an optional pinned THRML backend. Research snapshots and validation records are available in the public research repository. Full reproduction also uses the separately stored checkpoints and arrays.

\section{Discussion}
PBit1 links reconstruction fidelity to two configuration choices. Shared ranges set the scale of the error, and temporal allocation determines how much sampling is spent reducing it. The GPT-2 results show the impact of both choices at network scale. The head intervention and fixed-activation variance calculation identify a sensitive operator, although the layout change affects encoding bias as well as sampling variance. The native K4 result adds a spatial tradeoff, achieving lower operator error at matched scheduled observations through additional elements and coarse information.

The evaluation covers three checkpoints on two previously inspected tasks. Its five-seed intervals describe reconstruction variability on those fixed datasets, and the focused diagnostics use limited training inputs. Extending the study to untouched tasks, long-context generation, and instruction-following would establish how broadly the selected configurations transfer. Within the finite-control format, additional sampling improves estimation around the programmed mean, while command precision determines the remaining encoding bias. The supplement develops this distinction through a collision example and a counting bound.

A frozen-configuration sampling-budget sweep would map the quality--observation relationship more fully. Comparisons with optimized post-training quantization would connect representation fidelity to deployment costs. Full-model native execution and physical measurements would then extend the current BF16 reconstruction and THRML operator results to complete stochastic implementations, including programming, readout, retention, calibration, and correlated sampling.

\section{Conclusion}
PBit1 introduces and validates a concrete finite-control representation of pretrained neural networks, with one logical programmable p-bit per learned scalar, shared affine ranges, and temporal reconstruction. Shared ranges set the reconstruction scale, while temporal observations set estimation precision. The complete-model results establish that selected configurations preserve predefined quality margins on the evaluated checkpoints and tasks without fine-tuning. Sensitivity-based allocation shows that the temporal budget can be reduced while retaining that criterion.

Within the demonstrated regime, PBit1 establishes a finite-control, all-parameter path from pretrained weights to stochastic reconstruction, with measurable spatial, temporal, and fidelity tradeoffs.

\paragraph{Code, data, and declarations}
Package: \url{https://github.com/oramada/pbit1}. Research code and experimental records: \url{https://github.com/oramada/pbit1-research}. The manuscript and supplement are included in this package. The original project code is Apache-2.0; third-party model and dataset licenses remain separate. The THRML experiments use Extropic software in simulation. This work is independent of Extropic. See the companion supplement for evidence identifiers, exact protocols, and historical failures.

\balance
{\raggedright
\bibliographystyle{abbrvnat}
\small
\setlength{\bibsep}{2pt}
\interlinepenalty=10000
\bibliography{references-print}
}
\bigskip
\noindent\textbf{AI-assisted writing disclosure.} ChatGPT assisted with language editing, manuscript organization, reference discovery, and LaTeX formatting. The author takes responsibility for the final manuscript and its scientific content.
\end{document}


\maketitle
\tableofcontents

\section{Scope and reading guide}
This supplement accompanies the PBit1 finite-control representation manuscript. It records exact execution conventions, diagnostic controls, resource accounting, and the relationship between historical and accepted experiments. The original research plan is preserved as \code{master_design.tex}; it is a plan, not evidence that every proposed phase was completed. The research repository preserves completed, failed, interrupted, and unexecuted protocols separately. Only completed and audited observations support the numerical claims below.

There are three evidence levels: ideal independent-count simulation at full-model scale; direct official THRML sampling at operator scale; and physical devices. The third level is not part of this study. The term K1 counts one logical programmable stochastic element per scalar. It does not mean one bit of total storage, a physical allocation on a chip, or deterministic one-bit neural arithmetic.

\section{Checkpoint, task, and numerical contracts}
The evaluated checkpoints are GPT-2, \code{distilbert/distilbert-base-uncased-finetuned-sst-2-english}, and \code{LiquidAI/LFM3-1.7B}. LFM3 is pinned to revision \code{ac2586a34f1a1c03329dbdcdefabd8b6478ddaa7}. Its weight file SHA-256 is
\begin{quote}\small\nolinkurl{135922caef64e81e52af88ede8b7c3ddaa9d6da220a9737fb3065275dbfb4330}.\end{quote}
All remaining checkpoint and input hashes are in the bundled protocol files and research archive. Checkpoint weights are not redistributed in this Git repository.

Each R1 condition reconstructs one complete set of model weights and reuses it across all examples. Thus sampling seeds, not individual tokens, are the independent Monte Carlo units. GPT-2 scores 283,225 next-token targets over 586 windows and 62 articles. LFM3 scores 294,442 targets over 607 windows and 62 articles using its own tokenizer. DistilBERT scores all 25,000 balanced IMDb test reviews, truncated/padded to length 256, in batches of 32. The different tokenizer counts are intentional. These language-model evaluations are teacher-forced NLL measurements, not autoregressive generation benchmarks.

The original weights are rounded to BF16 before the stochastic comparison. Main model evaluations use BF16 neural computation. Diagnostic CPU experiments explicitly identified as such instead use FP32 arithmetic on BF16-rounded original and reconstructed weights. Eager attention and deterministic Torch settings are recorded in the source. LFM3 uses its original full-sequence output head and no KV cache. The custom fast convolution path is disabled in the recorded offload runs; this implementation choice is shared with their controls.

Native THRML is pinned to commit \code{2d54cd8404df85bb65703d7e8b4e84e48c1b72a9}, reported as version 0.1.4, with JAX 0.12.0.1 in the tested environment. Full-model runs use Torch 2.14.1+cu130 and NumPy 2.5.3. The optional SDK dependency specification records the complete tested pins. Runtime records are in the archive, rather than inferred from a package version alone.

\section{Finite encoder and reconstruction details}
\subsection{Grouping and constant groups}
The encoder accepts finite FP32-representable weights. A matrix can be flattened in storage order or transposed before flattening; the inverse permutation is applied after reconstruction. Groups are contiguous after that permutation, including a possibly shorter final group. A group stores its FP32 minimum and FP32 span, and one 12-bit command per scalar in a uint16 array. A zero span means the reconstructed group is exactly constant irrespective of the sampled state.

The probability field is clipped to $[-4,4]$ and rounded to 4,096 equally spaced command levels. Decoding evaluates the field in FP32 and the logistic probability in FP64. This exact order is important for hash replay. The supported group sizes in the SDK range from one through 65,536; the reported model configurations use 7, 16, or 128 as specified, not a per-parameter range that hides an extra full-precision copy.

\subsection{Independent counts and random keys}
Reconstruction calls NumPy's binomial generator once per scalar instead of emitting $T$ explicit Bernoulli bits. Each canonical tensor uses a random stream initialized from the experimental seed and tensor index. Groups are processed in deterministic chunks aligned to their boundaries. Counts use uint16 through $T=65{,}535$, and uint32 above that threshold. A SHA-256 digest over counts and a digest over reconstructed BF16 values are preserved in per-tensor records.

A binomial draw is an exact distributional shortcut only for independent identically distributed observations. It does not model temporal or cross-device correlation. The native and nonideal experiments are separate. Python random-number generation time also cannot be interpreted as a physical device's sampling time.

\subsection{API example}
The standalone package can reconstruct a finite-control program without model checkpoints:
\begin{verbatim}
import numpy as np
from pbit1 import FinitePBitProgram

w = np.array([[0.1, -0.2], [0.6, 0.3]], dtype=np.float32)
program = FinitePBitProgram.encode(w, group_size=2)
sample = program.reconstruct(32768, seed=7)
print(sample.weights)
print(program.storage())
\end{verbatim}
The return value is FP32. The research pipeline applies its explicit BF16 conversion afterward. The SDK's storage report counts array payload, not Python-object overhead, a reconstructed matrix, a retained reference model, or external hardware.

\subsection{Information bound}
For $T$ independent Bernoulli observations with interior probability $p$, the Fisher information is $\mathcal I(p)=T/[p(1-p)]$. The sample mean attains the Cram\'er--Rao variance bound $p(1-p)/T$ for unbiased estimation of $p$. Under a fixed affine range with span $d$, weight variance is $d^2p(1-p)/T$. This quantifies an estimation limit; it does not prove that a neural task needs that much parameterwise accuracy. Finite probability controls add a separate bias term that does not vanish with $T$.

\subsection{Why universal BF16 equality is excluded}
Two distinct BF16 values, $2^{-20}$ and $2^{-19}$, map to the same finite program in a tested $[0,1]$ range. No number of samples can distinguish identical programs. A counting argument gives a broader limit: with fixed groups of 128, the stated controls and ranges admit at most $2^{12.5N}$ programs for $N$ scalars. Even adding $N$ persistent binary states yields only $2^{13.5N}$ possibilities, fewer than the $65{,}280^N$ tensors of finite BF16 bit patterns. This argument applies to the stated format and fixed grouping; any extra metadata must also be counted. It excludes universal lossless conversion, while leaving model-specific approximation possible.

\section{Inference statistics and complete cohorts}
Table~\ref{tab:seeds} preserves all five selected-policy language-model perplexities. Their geometric mean corresponds to exponentiating average seed NLL; their arithmetic mean is not the reported main PPL. Seed IDs are experimental identifiers, not calendar dates.
\begin{table}[htbp]\centering\small
\caption{Complete language-model selected-policy cohorts.}\label{tab:seeds}
\begin{tabular}{rrrr}\toprule
GPT-2 seed & PPL & LFM3 seed & PPL\\\midrule
2026110801 & 25.60020 & 2026111901 & 18.98507\\
2026110802 & 25.54822 & 2026111902 & 18.59381\\
2026110803 & 25.58146 & 2026111903 & 19.17330\\
2026110804 & 25.92167 & 2026111904 & 19.01270\\
2026110805 & 25.69791 & 2026111905 & 19.13664\\\bottomrule
\end{tabular}
\end{table}

All primary intervals use the five seed-level deltas and a $t_4$ critical value of 2.776445. The final LFM3 joint seed/article bootstrap uses 5,000 replicates and produces [0.99188, 1.01065]. Recomputing statistics on saved target losses verifies coverage and scoring, but it does not establish independence of tokens or newly unseen data. Both are explicit limitations.

\subsection{Additional exact task and head-diagnostic values}
DistilBERT's mean accuracy is 87.4944\%, compared with 87.512\% for BF16. The upper 95\% interval endpoint for accuracy decline is 0.08383 percentage points; the categorical-perplexity ratio is 0.99966 [0.99676, 1.00256]. The reduced-work GPT-2 policy saves 41.676\% of nominal observations.

The fixed GPT-2 training diagnostic yields PPL 45.7987 for BF16 and 58.9795 for the original stochastic layout. Restoring only the input-embedding use gives 59.6616; restoring only the output-head use gives 46.1026; restoring both gives 46.2131. Reorienting the shared stochastic matrix gives 45.2709 in this draw. Under fixed clean hidden activations, predicted head logit variance falls from 0.61444 to 0.002698.

\figfull{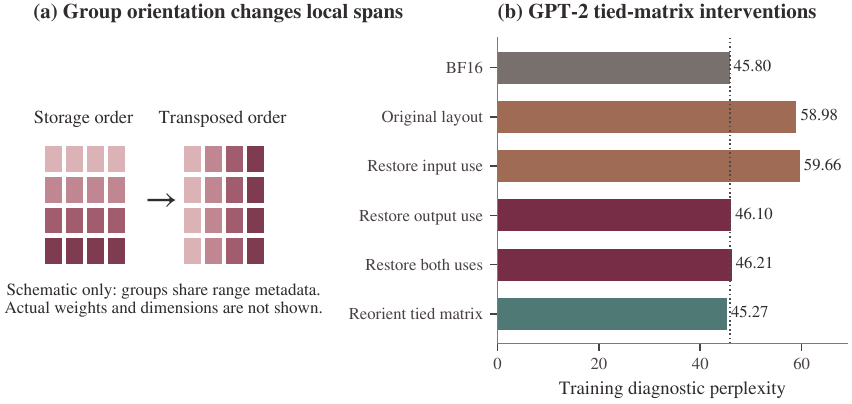}{1}{GPT-2 range-layout case study. Left: schematic grouping orientations; colors indicate shared ranges. Right: fixed-draw training interventions. Restoring the output use largely removes the degradation; the complete selected configuration instead retains one shared stochastic matrix. These diagnostics support a local variance explanation but do not separate the effect of changed encoding bias.}{fig:head_case}

\section{Exact sampling-allocation procedure}
The gradient proxy averages $\sum_i (\partial\mathcal L/\partial w_i)^2v_i$ over eight short training windows, where $v_i$ is the single-sample variance. Candidate one-dimensional ranges use group sizes 16, 32, 64, and 128. Matrix candidates retain group size 128 with the two flattening orientations. A second group of eight calibration windows evaluates proposals before freezing the complete GPU test.

The executed discrete allocation starts each tensor at $T_i=1{,}024$. Let $S_i$ be its selected gradient-weighted variance coefficient and $N_i$ its size. Doubling $T_i$ reduces proxy variance by $S_i/(2T_i)$ at an added cost of $N_iT_i$ observations. The greedy ratio is therefore
\begin{equation}
 \frac{S_i}{2N_iT_i^2}.
\end{equation}
The algorithm repeatedly doubles the eligible tensor with the largest ratio until it meets the recorded proxy target or reaches the cap $T_i\leq32{,}768$. The final accepted policy is \code{gradient_proxy_directions_vectorG16}. This is a fixed per-tensor schedule selected offline, not confidence-based early stopping for individual examples. The original plan's adaptive stopping and probability-aware fine-tuning remain separate possible extensions.

The full evaluation uses five fresh seeds, 2026111401 through 2026111405. Its frozen acceptance conditions require complete target coverage, all learned parameters K1, maximum sample count 32,768, nominal cost at most 60\% of uniform, and upper PPL-ratio interval at most 1.02. The observed cost is 58.324\%, with upper interval 1.01660. The BF16 reference arrays match the earlier complete baseline.

\section{Cumulative and causal diagnostics}
\subsection{GPT-2 cumulative substitutions}
The 41-condition CPU experiment consists of one baseline and eight configurations for each of five seeds: prefix and suffix groups of 3, 6, or 9 blocks, all twelve blocks, and all parameters. The arithmetic is FP32 on BF16-rounded weights, using 1,024 training targets. Nonblock parameters remain deterministic in the prefix/suffix and all-block controls; only the all-parameter condition converts every learned scalar.

\figfull{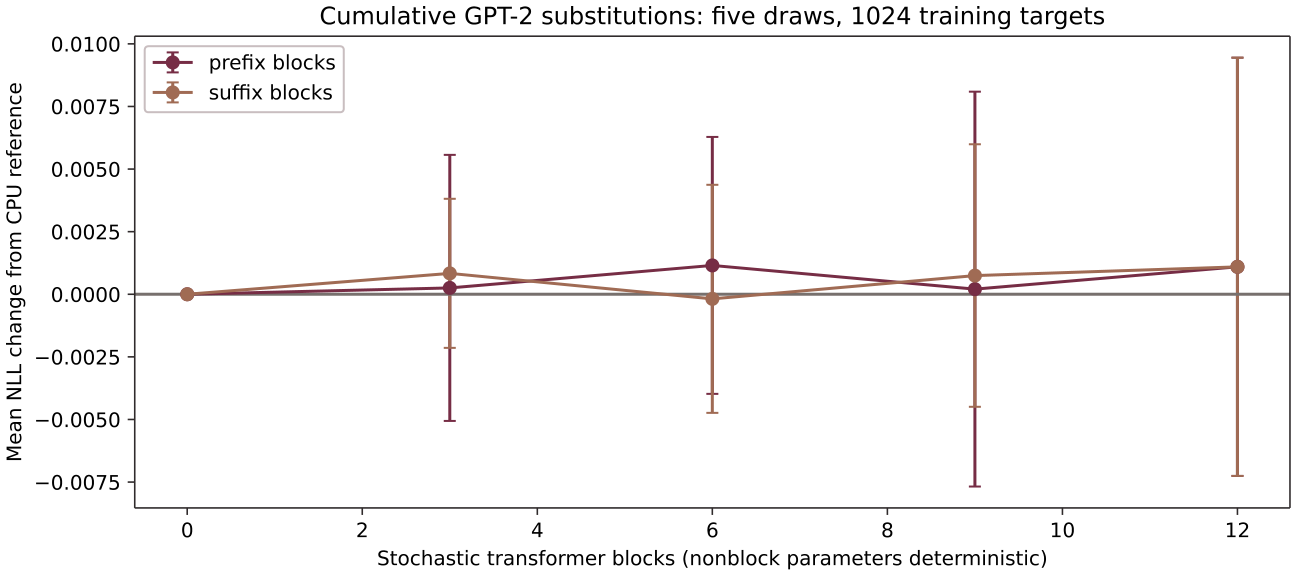}{1}{GPT-2 cumulative conversion on five training draws. Intervals overlap zero for the reported body-only curves, and loss is not monotonic in converted block count. These results do not prove a monotonic or additive layer-sensitivity model.}{fig:cumulative}

All masks, scores, and saved input contracts were audited; 148 SDK count hashes and six neural rows replayed. Small differences in loss must be interpreted at this limited sample size, rather than as a ranking of universally dispensable layers.

\subsection{LFM3 intervention coverage}
The 55-condition LFM3 causal study uses eight training windows of 129 input tokens, yielding 1,024 scored targets. The fixed draw uses seed 2026111801. One baseline, one all-K1 condition, 24 isolated substitutions, 24 restorations, and five family restorations are recorded. The isolated interventions leave every earlier block unchanged. Six independently reconstructed neural training rows and all 286 count records establish the replay contract before the intervention analysis.

\figfull{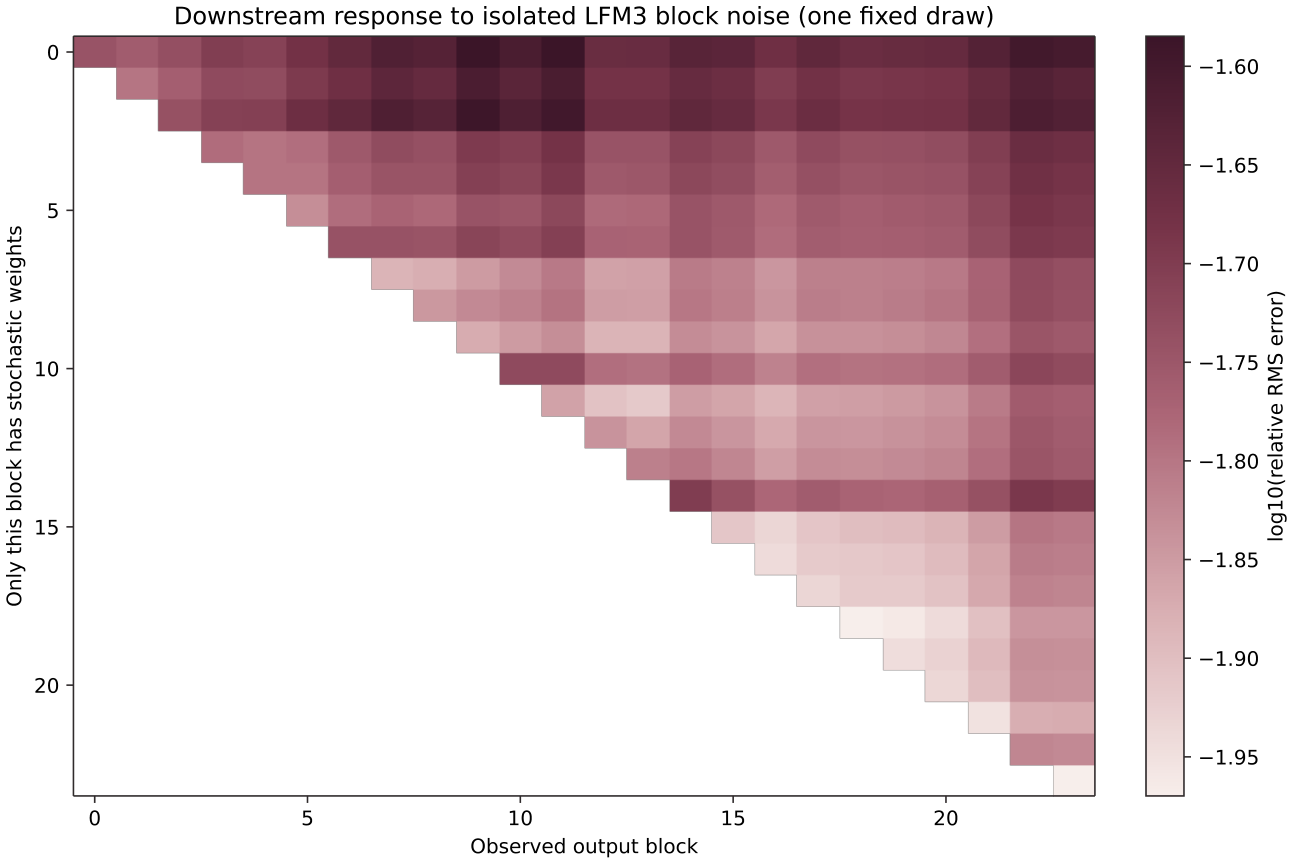}{.94}{Downstream LFM3 responses to isolated substitutions. Earlier blocks are exactly unaffected. Absolute responses depend on the operating point and include subsequent nonlinear computation; this matrix is not a global stability guarantee.}{fig:propagation}

In the fixed draw, block 3 is the most harmful isolated substitution, whereas restoring block 2 gives the largest improvement in the fully stochastic network. The differing rankings show that sensitivity depends on perturbations elsewhere in the model.

\figfull{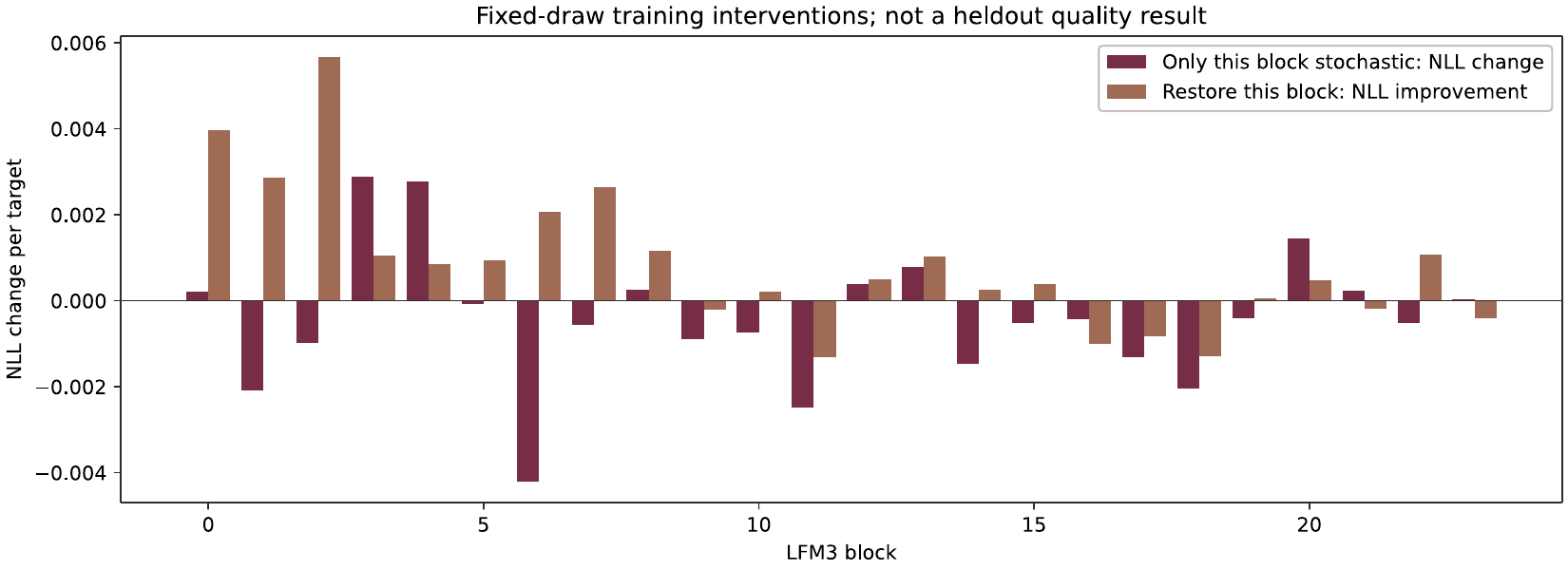}{1}{LFM3 isolated-block and restoration diagnostics on 1,024 training targets and one fixed weight draw. Layer numbers are zero-based. The interventions distinguish isolated sensitivity from effects in the already-perturbed network.}{fig:causal}

\subsection{Signal scale and neuron structure}
Between LFM3 blocks 11 and 12, error power rises 20.76\%, but signal power rises 57.03\%, causing relative RMS error to fall 12.305\%. The relative improvement therefore does not imply denoising.

For each LFM3 block, the baseline activation study uses 8,224 positions and 2,048 channels. The activation energy participation ratio is $(\sum_j e_j)^2/\sum_j e_j^2$, where $e_j$ is the channel second moment. Ratios range approximately from 9 to 654 across blocks. This measures concentration of activation energy, not the number of removable neurons.

\figfull{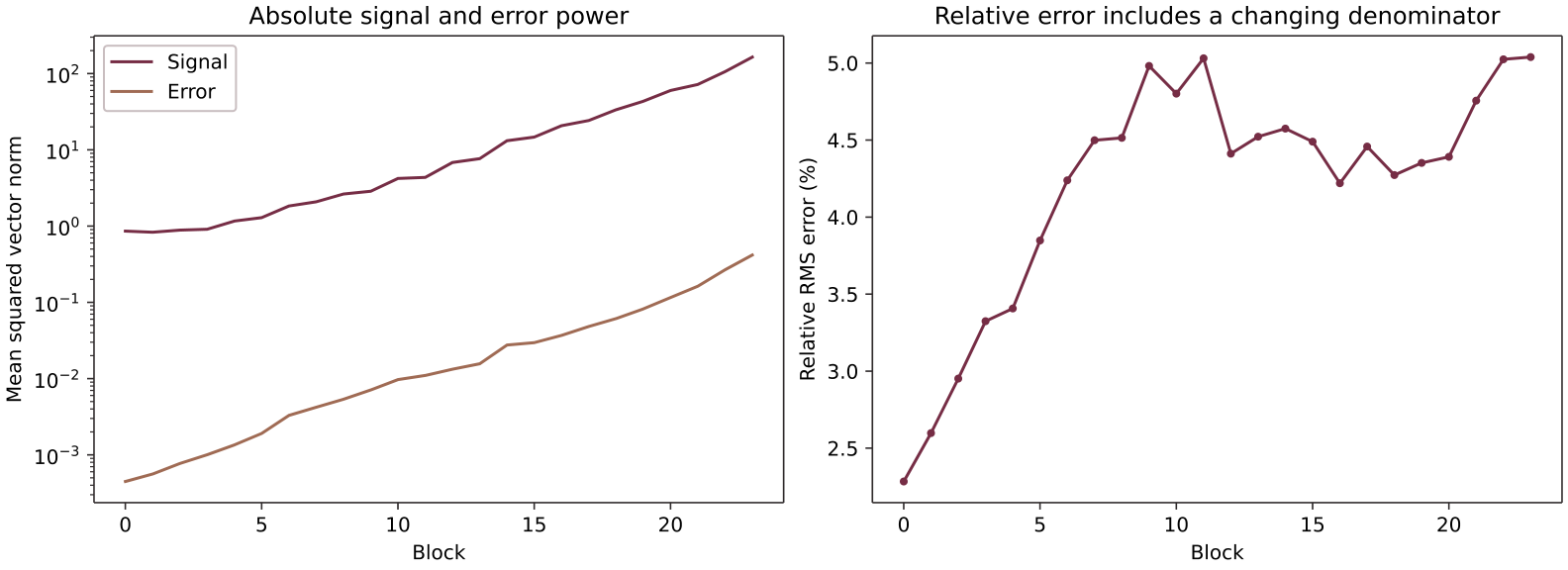}{1}{LFM3 absolute error, signal scale, and relative RMS error across blocks. A decline in relative error can occur while absolute error rises; the curves alone do not isolate denoising.}{fig:scales}

\figfull{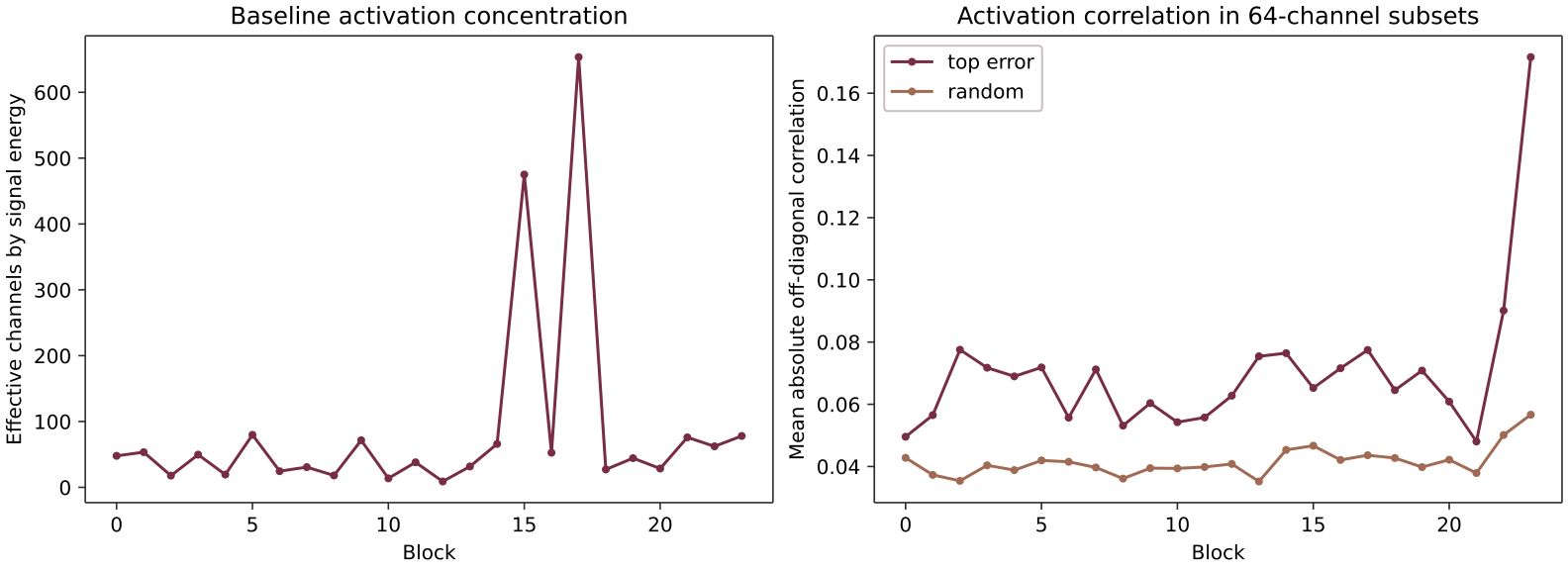}{1}{Baseline LFM3 channel structure and error concentration. Correlations of baseline activations in this figure are distinct from the conditional stochastic-error covariance in the main manuscript.}{fig:neurons}

\section{Covariance estimator and interpretation}
Let $e_{s,t}\in\mathbb R^{64}$ denote the output error for seed $s$ at fixed position $t$, relative to the deterministic baseline, and let $\bar e_t$ be its mean across $S=5$ seeds. The estimator is
\begin{equation}
 \widehat C=\frac1M\sum_{t=1}^{M}\frac1{S-1}
 \sum_{s=1}^{S}(e_{s,t}-\bar e_t)(e_{s,t}-\bar e_t)^\top,
 \qquad M=8{,}224.
\end{equation}
It averages conditional covariance over input positions. It is not obtained by pooling positions and subtracting one global mean. Sufficient statistics include cross sums, the mean outer-product contribution, channel indices, and the resulting matrices. The independent auditor checks all 48 matrices and a direct small-array covariance fixture. All five model draws replay all 8,192 training target losses bitwise before covariance accumulation.

The participation ratio for covariance is $(\operatorname{tr}C)^2/\operatorname{tr}(C^2)$; the leading-eigenvalue fraction is $\lambda_{\max}(C)/\operatorname{tr}C$. The high-error channels were selected using this same training cohort. Their more concentrated covariance therefore cannot be interpreted as an unbiased population comparison. Averaging conditional covariances may have rank greater than $S-1$ because input positions have different error subspaces.

Mean absolute off-diagonal correlation ranges from 0.0246 to 0.1066 across blocks in the high-error subset, compared with 0.0166 to 0.0327 in the random subset. In the last block, the high-error subset has covariance participation ratio 4.50 and leading-eigenvalue fraction 0.460; the random subset has participation ratio 56.55. Independent weight sampling can therefore produce correlated output errors through shared computation. These statistics describe the selected 64-channel subsets rather than the full 2,048-channel covariance.

\figfull{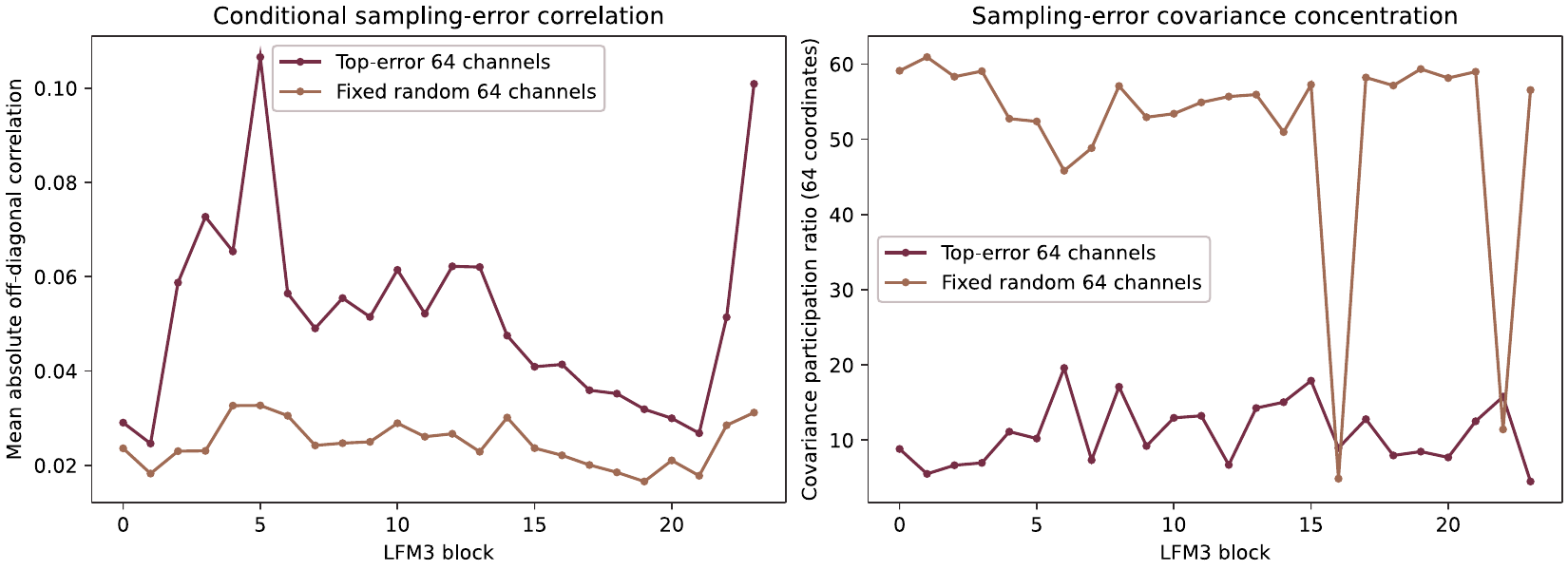}{1}{LFM3 conditional output-error covariance summaries. Covariance is computed across five draws at each fixed input position and averaged over positions. High-error channels are selected on the same cohort, so differences from the random subset are descriptive.}{fig:covariance}

\section{Additional controls and historical outcomes}
\subsection{Matched Gaussian noise}
Figure~\ref{fig:gaussian} presents the training-only, full-parameter GPT-2 comparison. The noise means and per-coordinate variances are matched before FP32/BF16 rounding. All eleven score cases and 740 paired variance entries were audited, and three Gaussian parameter reconstructions independently replayed. The two cohorts use independent random keys. Overlapping intervals do not prove equivalence; they show that these observations do not isolate a special Bernoulli quality benefit.

\figfull{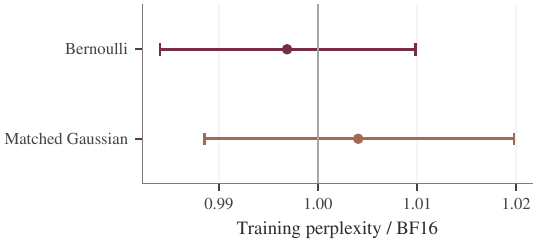}{.96}{Full-model training diagnostic with matched Gaussian and Bernoulli estimator variance. These are independent five-seed cohorts on fixed training texts, not additional complete held-out results.}{fig:gaussian}

\subsection{Correlation, drift, and calibration}
Operator-level moment calculations and simulated calibration probes examine departures from ideal independent sampling; these nonideal models were not executed through native THRML. A stationary binary process with lag correlation $\rho^k$ has finite-sample variance inflation
\begin{equation}
 1+2\sum_{k=1}^{T-1}(1-k/T)\rho^k,
\end{equation}
which approaches $(1+\rho)/(1-\rho)$ for large $T$. Accordingly, the common approximation $T_{\rm eff}\approx T(1-\rho)/(1+\rho)$ has an explicit finite-$T$ limitation. Cross-device correlation adds covariance terms to output variance and must not be represented solely by a scalar effective sample size.

The preserved head investigations include 900 nonideality cases, and finite-probe, prior-aware calibration analyses with 140 cases and 48 optimizer checks. Their calibration priors are assumptions, not measured device populations. An unregularized calibration fit can worsen estimation; an appropriate prior can help in the tested simulator. Probe counts and calibration storage belong in the cost ledger. Drift sensitivity is simulated; no hardware recalibration interval is established.

The gain/offset model is
\begin{equation}
 p_i^{\rm dev}=\sigmoid\{a_i\operatorname{logit}(\tilde p_i)+b_i\},\qquad
 a_i\sim\mathcal N(1,0.02^2),\quad b_i\sim\mathcal N(0,0.02^2).
\end{equation}
On 1,024 training targets, the paired uncalibrated-to-ideal PPL ratio is 1.03229 [1.00038, 1.06522]. An oracle with exact knowledge of each gain and offset reduces mean squared bias by about 447-fold, with clipping and control quantization remaining. Two FP32 calibration values per GPT-2 scalar require 995,518,464 bytes. This is an idealized correction experiment, not a demonstrated device-calibration procedure.

The full GPT-2 gain/offset experiment uses 16 CPU training cases: one baseline and five seeds each for ideal, mismatched, and oracle-corrected conditions. It audits the matched device arrays and nine parameter replays. Oracle coefficients are known exactly, then re-quantized to the finite control grid. This separates programming bias from sampling variance but does not solve the physical calibration problem.

\subsection{Earlier experiments and failures}
The earlier adapted LFM3 checkpoint study used 2,048 fixed validation targets and a same-fit BF16 reference of PPL 75.3147. Its T1024 cohort gave PPL 83.9167 and missed the 2\% criterion; T4096 gave 76.1024 and also missed the criterion because its upper interval exceeded the margin. T8192 gave 75.1547, ratio 0.99788 with interval [0.98062, 1.01544], and passed. These adapted, shorter-task results are not pooled with the final original-checkpoint evaluation.

An earlier GPT-2 T8192 cohort failed, whereas DistilBERT passed its original criterion. The local range diagnosis in the main paper investigates this variation. An additional 10,240-target LFM3 cloud study was interrupted and has no accepted final quality outcome. A GB10/H100 bridge reproduced stochastic and nominal arrays, while its BF16 reference shifted approximately 0.00515\% in perplexity. The initial strict bitwise-control failure is preserved, together with the later documented hardware-specific-reference amendment. These observations are historical context, not additional independent proof of the final Spark claims.

The archive also preserves failed implementation and audit attempts, including calibration bookkeeping errors and revised auditors. Only the corrected completed runs are used for accepted statistics, and exact executed source snapshots remain available. Repeated audits of the same data are not counted as new experimental replications.

\section{Resource ledger and storage boundary}
\begin{table}[htbp]\centering\small
\caption{Exact array payload counts. Commands are the implemented uint16 arrays; ranges are two FP32 values per group. Decoded BF16 weights, conceptual binary states, software objects, and transient buffers are separate.}\label{tab:storage}
\begin{tabular}{lrrr}\toprule
Configuration & Range groups & Command bytes & Range bytes\\\midrule
GPT-2 uniform & 972,186 & 248,879,616 & 7,777,488\\
GPT-2 reduced & 978,822 & 248,879,616 & 7,830,576\\
DistilBERT & 523,087 & 133,910,020 & 4,184,696\\
LFM3 & 13,542,448 & 3,457,603,072 & 108,339,584\\\bottomrule
\end{tabular}
\end{table}

Ideal packed controls are 12 bits per scalar, but the current NumPy implementation stores them in uint16. No packed 12-bit runtime is claimed. Fixed shape, tensor addressing, layout flags, per-tensor sample budgets, and Python bookkeeping incur further overhead. There is no residual side table in the accepted K1 model configurations and no excluded deterministic learned subset. Diagnostic restoration controls explicitly relax that condition.

R1 reconstructs weights once for repeated inference. This preserves a decoded weight array. Discarding the original model after encoding can remove a reference copy, but not the working weights needed by this implementation. Device density, bit storage, free random transitions, all-node observations, and elapsed latency are separate metrics. The original four-element word baseline and the stronger coarse/fractional baseline differ in stored control information; neither licenses a blanket systems-efficiency statement.

\section{Implementation replay coverage}
Independent replay reconstructs twelve GPT-2 and twelve DistilBERT weight configurations and checks 36 neural windows or batches for each architecture. For LFM3, it checks seven configurations, 1,430 stochastic parameter-count hashes, and 21 full-network windows. Every prescribed comparison matches bitwise on the recorded runtime. These are implementation checks on selected inputs, not independent full-dataset evaluations.

The reduced-work validation additionally reports eighteen prescribed neural-window replays. Uniform and reduced policies use separate seed cohorts; meeting the same acceptance criterion does not establish identical quality. The savings count nominal Bernoulli observations for a reconstruction reused across inputs.

All thirty native cases are audited, and three full trajectories replay bitwise. Analytical expectations include cross-input covariance and fourth moments; a sixteen-outcome enumeration checks the moment formula. The reported development SDK distribution passes 23 installed-package tests without skips and a native CLI smoke test.

\section{Evidence, reproduction, and outstanding work}
\begin{table}[htbp]\centering\footnotesize
\caption{Selected accepted evidence identifiers, relative to the dated Spark result directory.}\label{tab:evidence}
\begin{tabularx}{\linewidth}{lX}\toprule
Claim & Artifact identifier\\\midrule
Three complete models & \code{cross_model_summary_v2/results.json}\\
GPT-2 saved-output audit & \code{benchmark_audit_v1/results.json}\\
DistilBERT saved-output audit & \code{distilbert_benchmark_audit_v1/results.json}\\
LFM3 saved-output audit/replay & \code{lfm3_benchmark_audit_v1/}, \code{lfm3_benchmark_replay_v1/}\\
Tied-head intervention & \code{tied_head_v1/results.json}\\
LFM3 causal analysis & \code{lfm3_causal_depth_analysis_v1/results.json}\\
Covariance audit & \code{lfm3_error_covariance_v2_audit.json}\\
Reduced-work audit/replay & \code{reduced_work_gpu_v1_audit.json}, \code{reduced_work_replay_v1/}\\
Strong native control & \code{native_coarse_fractional_v1_audit.json}\\
Resource and precision limits & \code{resource_ledger_final_v1.json}, \code{precision_limits_v1/}\\
SDK tests and package hashes & \code{sdk_build_verification_v2.json}\\\bottomrule
\end{tabularx}
\end{table}

The package and bounded archive preserve source snapshots indexed by SHA-256, protocols, summaries, independent audits, package versions, and omitted-asset hashes. The manuscript's \code{EVIDENCE_MANIFEST.json} maps each bundled figure/input to its originating artifact. Manuscript plots derived from JSON can be regenerated without model weights; raw-output auditors require the omitted model assets and arrays or complete reruns. A hash manifest verifies integrity, not scientific correctness by itself.

For model reproduction, first restore the pinned checkpoints and dataset layouts under their original licenses, recreate the recorded runtime, and use a new output directory. Producers preserve original protocols and typically reject conflicting or already-failed state. Do not edit executed producer files to force old results to pass. Large-model replays, native samplers, and CPU diagnostics use different runtime environments as recorded. Representative module commands are documented in the repository reproduction guide.